# Data Interpretation Support in Rescue Operations: Application for French Firefighters

Samer CHEHADE*†, Nada MATTA*, Jean-Baptiste POTHIN†, Remi COGRANNE*

*Insitut Charles Delaunay, TechCICO / LM2S*
*Université de Technologie de Troyes*
Troyes, France
{nada.matta, remi.cogranne}@utt.fr

† *Department of Research & Development*
*DataHertz*
Troyes, France
{samer.chehade, jean-baptiste.pothin}@datahertz.fr

*Abstract*—This work aims at developing a system that supports French firefighters in data interpretation during rescue operations. An application ontology is proposed based on existing crisis management ones and operational expertise collection. After that, a knowledge-based system will be developed and integrated in firefighters' environment. Our first studies are shown in this paper.

## I. INTRODUCTION

Rescue of people consists in saving their life in case of distress situations by applying responsive operations. In France, it is defined as specific tasks to be accomplished by public services in order to ensure the safety of patients and victims by making them able to escape from dangers, securing intervention sites, providing medical help, and finally, ensuring the evacuation to an appropriate place of reception [1]. These services have to deal with several flows of information coming from different sources. They have to make a decision based mainly on these data. Semantic representation techniques [2] [3] can be very helpful for them in order to make the right decision at the right moment. In this paper, we propose to build a dedicated ontology to represent French firefighters' activity as well as information and data flow in rescue operations.

This paper is divided in three sections:

1. Definition of terms used in rescue operations and an illustration of data flow complexity
2. Semantic representation of data and information and related existing ontologies
3. ResOnt, the ontology we defined to help French firefighters

## II. RESCUE OF PEOPLE IN DISTRESS

### A. Definitions

Rescue of people, also named emergency rescue of people is one mission of firefighters and medical units. It consists in ensuring the safety of patients or victims by making them able to escape from a danger, securing sites on interventions, practicing first aid gestures, and, finally, ensuring the evacuation to a suitable place of reception.

In France, two public services are engaged in rescue operations and emergency care to the population: Departmental Fire and Rescue Service (SDIS) and Emergency Medical Assistance Services (SAMU) [4]. SAMU are responsible of ensuring medical assistance while SDIS have a lot of tasks and missions including medical care in some cases. these two services have to cooperate and each one should be aware of the tasks and mission ensured by the other. In order to delimit missions and make responsibilities more clear, a common referential has been elaborated [1]. This referential focuses on the sharing of responsibilities as well as making distinctions between duties of each service involved in rescue operations [1]. The mission of SAMU is limited in [1] to providing medical help in emergency situations. This task can be divided into the following five subtasks: 1) Provide permanent medical listening, 2) determine and trigger best responses to the nature of the calls, 3) ensure the availability of the means of hospitalization adapted to the patient's or victim's condition, 4) organize the transportation of the victim using a public service or a private health transport company and, 5) ensure the patient's admission [1]. Whereas the mission of SDIS is to prevent, protect and fight fires [1]. They are also responsible of other missions, in particular the rescue of patients or victims of accidents or disasters as well as their evacuation [5]. Some other services and centers participate indirectly in rescue operations and emergency responses. These services and centers are connected to the two previous ones and are responsible of receiving calls, transferring calls to the good actor as well as ensuring a good communication and information exchange between SAMU and SDIS during and after a rescue operation. Each SDIS has a fire and rescue departmental operational center (CODIS) responsible for coordinating the operational activity of fire and rescue services at the departmental level. In addition, it has one or, if necessary, several alert processing centers (CTA), responsible for receiving, processing and possibly reorienting relief requests. Moreover, it possesses health and medical rescue service that participates, in particular, in emergency relief missions (SSSM) [1]. On the other hand, each SAMU is connected to the Center of Receiving and Regulating Calls (CRRA) responsible of receiving calls and transferring them to the CTA of SDIS when the case imposes the participation of this latter in a rescue operation. They are also connected to Mobile Service of Emergency and Reanimation (SMUR) that provides care for a patient whose condition requires, urgently, medical care and resuscitation.

## B. Information flow and complexity

Different rescue services, whose common objective is to save lives, handle different tasks of the job. It is therefore critical for them to cooperate, communicate and coordinate. Add to that, information related to operations process and victim's status should be shared between actors as soon as possible. Information is the most important resource for coping and recovery work in rescue operations. It is the foundation for coordination and decision-making in best delays. Information take several forms such as text, audio and video. An example of information to be shared during a rescue operation is rescue record done by firefighters on the ground and sent to SAMU in order to prepare for the most suitable medical response to the victim. In order to ensure a good communication and sharing of information, [1] imposes the interoperability between CRRA and CTA. The interconnection requirements are of two kinds: the first one is for audio communications (through public telephone lines and radio communications systems). The second one is for data through ANTARES network (National Adaptation of Risk and Rescue Transmissions). Due to the complexity and diversity of information, rescue actors usually have a problem to identify critical events to deal with at the right time in order to reduce consequences of urgent state. These actors have to make critical decisions in a very short lapse of time; each second can be important.

The objective of this work is to use semantic representation techniques in order to help rescue actors to understand more quickly critical events. From this point, we propose a new application ontology for rescue operations that will help in automating data processing and saving a lot of time.

## C. Related work

During the last years, emergency responses and rescue operations have merged technologies of communication, computer, artificial intelligence, knowledge engineering and system engineering. Several systems and applications have been developed over the last ten years in order to help decision makers to ensure quick responses in these operations such as ISyCri [6], SHARE [7][8], RESCUER [9], DISASTER [10] [11], PIECES [12]. These systems aim to provide rescuers with mobile technologies and services to support information management multimodal communications, interoperability of actors collaborative processes. More recent systems are oriented to decision-making in crisis situations, for example, Crisis Clever system [13], Decision support system for emergencies [14] and systems that focus on information flow and data processing during disasters such as $S^2D^2S$ [15].

As mentioned above, our aim in this paper is to build an ontology to support firefighters in data interpretation during and after a rescue operation. Gruber defined ontology as *"an explicit specification of conceptualization"* [16]. Bachimont, declared that defining an ontology for a given domain and problem, is defining the relational and functional signature of a formal representation language and the related semantics [17]. Ontologies can be classified into three categories, each one having its own specifications and use cases [18]. These categories depend on the level of generality of the ontology, and are separated into *top-level ontologies*, also called *upper ontologies*, *domain ontologies*, or *application ontologies*. Top-level ontologies are independent of a specific domain or problem. They define general concepts such as action, event, time, object, matter, time, space and many others that can be used in several domains. Examples of these ontologies are SUMO (The Suggested Upper Merged Ontology) considered as the largest accessible ontology [19], Sowa's Ontology represented in Sowa's diamond [20], OpenCyc ontology [21], DOLCE defined by its authors as an ontology of particulars [22], and BFO developed by the Institute for Formal Ontology and Medical Information Science (IFOMIS) [23]. Domain ontologies are specification of top-level ontologies. They are related to a generic domain such as crisis management or rescue operations. Domain ontologies can also be specialized in order to build application ontologies that describe specific concepts in a particular domain and particular associated tasks. Good examples of domain ontologies are SOKNOS [24], ontologies related to crisis management and response domain [25] [26] or ontologies for medicine and biomedical domains such as the Biodynamic Ontology which is a domain ontology based on the BFO upper ontology [23]. Moreover, many application ontologies have been built from which we cite EMERGEL ontology [10], EDXL-RESCUER [27] and SHARE-ODS [7] developed to support DISASTER [10] [11], RESCUER [9] and SHARE [7] [8] projects respectively.

Our aim is to construct an application ontology related to rescue operations and to be more specific, to the organization, responsibilities, communications, processes and flow of information during a rescue operation ensured by firefighters in order to support them in data interpretation. However, aspects of coordination and cooperation are not covered in our ontology. During our literature review, we found several ontologies developed in the context of rescue operations. However, these ontologies are not publicly accessible and not directly applicable in our context such as SHARE-ODS [7]. In addition, we also found several domain and application ontologies constructed in crisis and emergencies management domain to support decision makers. These ontologies can serve us to borrow ideas and some general concepts defined in these latter in order to construct our proper ontology specific for rescue operations ensured by French firefighters. However, they cannot be used completely in our case because a rescue response differs from crisis or emergency response in many concepts. In addition, some important concepts related to our context are not well defined in previous ontologies. For instance, actors organization and structures, dataflow, roles and many other concepts. From the ontologies that we found, we based our work on four ontologies that are the most suitable to our interests. These ontologies are EXDL-RESCUER [27], EMERGEL [10], the emergency ontology [25] and the emergency response ontology [26]. Thus, we will give more details about these ontologies and the case of use of each one in section III.A. The advantage of our ontology is that it defines new concepts related to rescue operations, which are not defined in other ontologies.

III. SEMANTIC REPRESENTATION OF DATA

The complexity to process data coming from a variety of sources, and the lack of time during a rescue operation necessitate automation of information management procedures. Our goal is to define a methodology to help actors involved in rescue operations in data interpretation. Our methodology is based on semantic data recognition. It consists in using techniques and methods able to represent the human's expertise in a certain domain in order to simulate problem solving and data recognition. One of these techniques is Knowledge engineering, which enables modeling expertise's concepts [2]. This type of representation involves the construction of ontologies. Sowa defined knowledge representation as the use of ontologies and logic to build computable models in a domain [3]. Concepts can be represented at two levels: as a semantic network close to conceptual graphs defined by Sowa [28] and in a computable way using logic languages [29]. Related to our purpose, several works have been distinguished in the context of knowledge engineering in order to build crisis and emergency response systems such as knowledge-based models proposed to support environmental emergency management [30], or models to solve unconventional emergencies problems by using domain knowledge and ontologies [31].

*A. Existing Ontologies*

In this section, we give details about ontologies in order to adapt an ontology corresponding to firefighters rescue operations.

The first ontology chosen in this paper is a 'Domain Ontology' developed to support knowledge reorganization in decision support systems [25]. Authors constructed this ontology based on emergencies documentation by adopting the Activity First Methodology [32]. They defined four main classes: **EVENT** that can be a *disaster* or a *disease*, **RESOURCE** made of *artificial* and *inartificial*, **SUBJECT** divided into *personal* and *actor*, and **TASKS** consisting of "communication", "evaluation", "rescue", "prevention", and "detection". Each one of these classes is also divided into different subclasses, each having several instances. In the domain of rescue operations, this ontology can serve by giving the general concepts that are in common with crisis management. This ontology has the advantages to cover different contexts in emergency response, ranging from events to tasks, passing by subjects.

The second ontology we studied is the 'Task Ontology' that was constructed to support the implementation of the evacuation planning system for emergency cases and to standardize a group of semantic concepts used in different emergency systems [26]. It defines a common vocabulary usable by emergency actors regardless of the emergency nature. To build this ontology, Xiang et al. identified four generic concepts that are Response Preparation, Emergency Response, Emergency Rescue, and Aftermath Handling. Each one of these concepts is divided into several sub concepts, illustrating the main steps and tasks to be taken in case of emergency. This ontology is restricted to the main missions to achieve during an emergency case from which we will take some concepts that are in common with rescue operations such as medical aid, communication and victim assistance.

The third ontology we consider is the EXDL-RESCUER ontology [27] created to support the RESCUER project [9], which uses crowdsourcing information to assist actors in emergencies. This application ontology is based on Emergency Data Exchange Language and aims to construct a conceptual model correspondent to information exchange and coordination with other systems. It focuses on the type of exchanged information and it covers a group of message contexts in an emergency. In this ontology, authors defined twelve concepts related to the type of shared messages during an emergency response such as Alert, Info, ResponseType, MsgType, etc.

Finally, we studied the EMERGEL ontology [10] constructed to support the DISASTER project [10] [11], which focuses on Data-Interchange on a semantic level. This application ontology is constructed to temporally describe a crisis situation by defining different modules divided into transversal modules and vertical modules. In order to construct this ontology, authors chose to specify some top-level classes defined in the generic ontology DOLCE [22]. Transversal modules are time and space while vertical modules are objects representing concepts in real word, constructs representing a set of objects, and activities representing tasks to be achieved by objects and constructs. Each one of these modules is divided into many classes. For example, objects contains Person, Vehicle, Equipment, Infrastructure, SpatialPoint and Communication. This ontology is various and cover different subjects in disaster response.

We can note that in these ontologies different aspects are represented related to different goals of research: studying tasks, communication, operations, etc. In our work, we focus on data interpretation, so communication aspects are important in our work. However, communication concepts must be linked to tasks and means. We present in the following how these concepts can be combined in an application ontology.

*B. ResOnt: Proposed Application Ontology*

In this section, we discuss the "ResOnt" ontology we constructed to support data interpretation in rescue operations. To construct this ontology, we took several concepts defined in the studied ontologies detailed in III.A. Add to that, we analysed documents related to rescue operations in France [1] [4] [5], and we conducted an interview with an expert Firefighter in Aube's Department. Then we identified new concepts, we reorganized them with the existing ones extracted from the studied ontologies, we completed them and we built the rescue operations ontology. First of all, we defined the rescue operation as a world and we divided it into two generic concepts **OBJECTS** and **CONSTRUCTS** as it has been done in EMERGEL [10] ontology in order to make a first classification between different categories. These two categories represent endurants and perdurants defined in DOLCE [22]. We chose to follow the classification of DOLCE [22] since it provides concepts that can be the basis for interoperability between lower level ontologies and it is used in multilingual information retrieval and web-based systems and services. Objects are defined as a representation of concepts in the real world. In our ontology, the main category **OBJECTS** includes three main concepts that are *Material*, *Person* and *Infrastructure* and each one is divided into several sub-concepts as shown in Fig. 1. In Material, we consider all concepts and instances that are in relation with inartificial resources used during a rescue operation including vehicles, catering, safety, medical, technical equipment and for sure the equipment used to ensure communications. In persons we consider all people involved in a rescue operation. It is divided into two parts: victims who are the object of an operation and actors who are the subject of an operation. While infrastructure represents a site intervention that could be a road, a bridge or a building. On the other hand, Constructs represent concepts that are different from touchable objects. This category includes five main concepts that are *Incident*, *Organization*, *Role*, *Tasks*, and *Data Flow* divided each into many sub-concepts as shown in Fig. 2. *Incident* represents the type event behind the intervention. It can be a disease, a natural event causing victims, a fire or an accident.

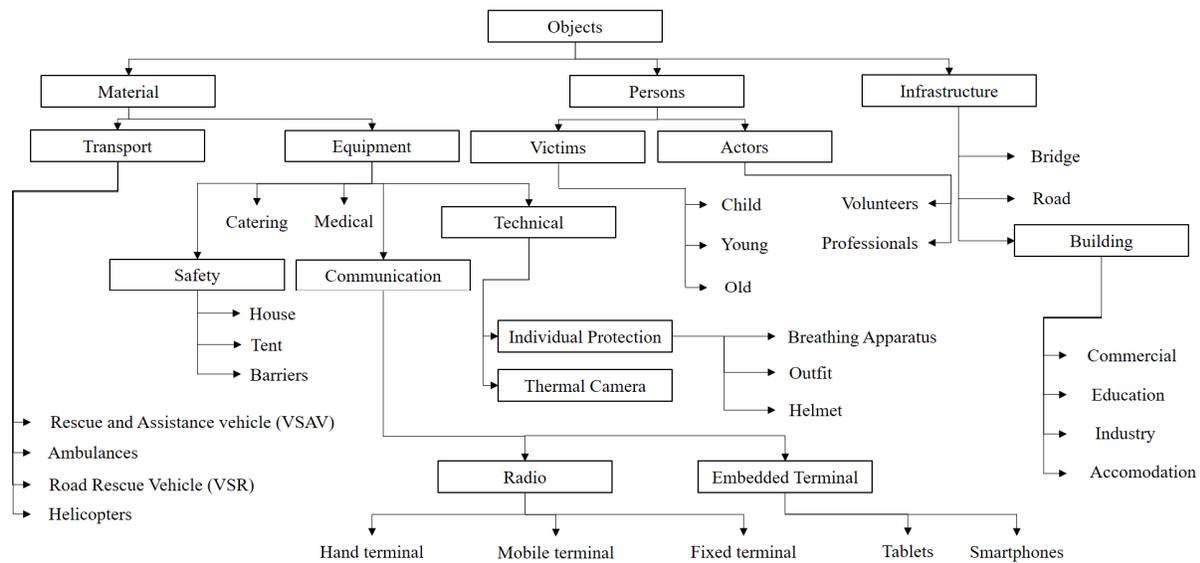

Fig. 1. Hierarchy of Objects

*Organizations* represent the main organisms that participate in rescue operations in France and contains Red Cross, SDIS and SAMU connected each to different centers and services while *Role* represents the order of each person participating in an operation. *Data Flow* represents the exchanged information during, before and after an operation. It contains information about the weather, actors, the evolution of an operation and the situation during an operation. And finally, *Tasks* represents duties and missions needed to ensure a successful rescue operation. These tasks covers four main phases : 1) Preparation including all tasks to be taken before an operation in order to achieve any operation without wasting time, 2) Response including Coordination, Communication and Decision-Making, 3) Operation including tasks to be taken on the intervention site and, 4) tasks to be taken at the end of the operation such as reporting, investigation and social assistance of victims. However, due to a lack of space and the big number of concepts defined in tasks, we decided to represent them in a separated figure instead of putting them in the same figure of constructs as shown in Fig. 3.

The work done represents our first step in the implementation of the ontology. In this step, we defined concepts used in rescue operations and we made a first classification in a hierarchical tree. However, relations between these concepts have not been identified yet. Once done, we look forward implement, evaluate and submit it to an ontology catalog.

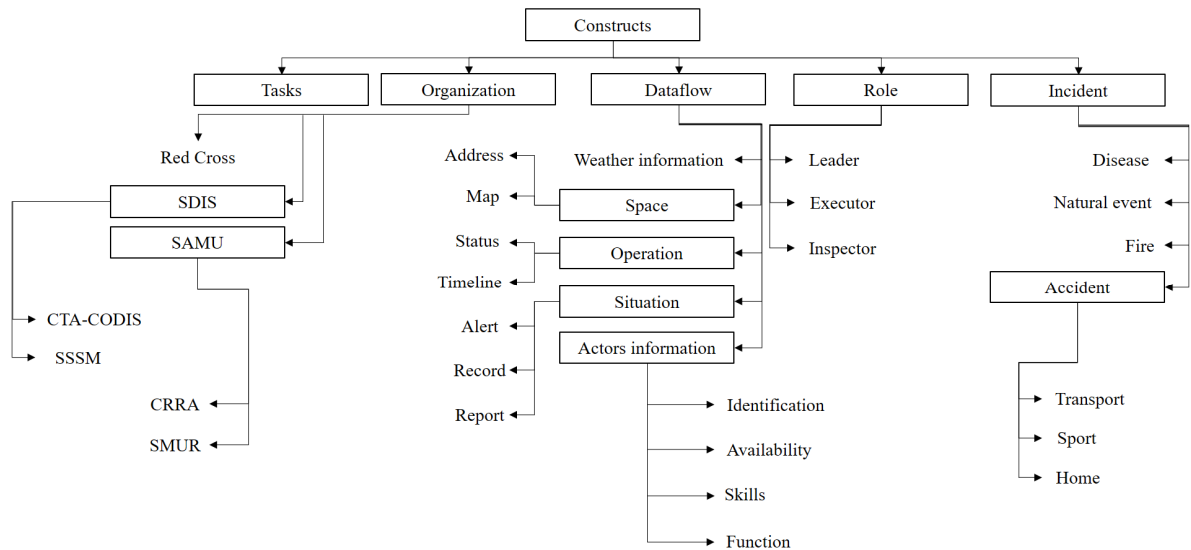

Fig. 2. Hierarchy of Constructs

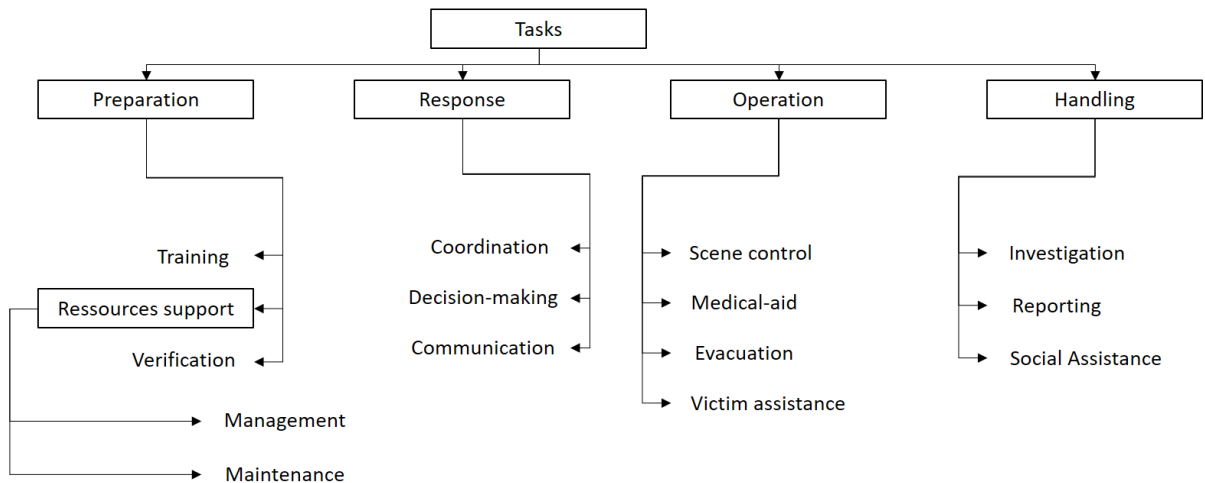

Fig. 3. Tasks to be achieved

IV. CONCLUSION

In this paper, we presented the construction of an application ontology dedicated to French firefighters participating in rescue operations after studying several ontologies constructed in crisis management domain. This ontology will be the base of a system aiming to support firefighters in data interpretation. However, our ontology is not evaluated yet since it is our first study in this aim and the ontology is not implemented.

As a future work, we will detail this ontology and instantiate it based on expertise collection and modelling. Thus we will conduct series of interviews with firefighters operational staff involved in rescue operations. After that, we will implement our ontology in order to test it in real cases ,evaluate it and submit it to an ontology catalog. Finally, we will study the design and implementation of an information recognition system based on knowledge bases by exploiting the ResOnt ontology. This system will be integrated as a decision-support system in French firefighters working environment. This work includes naturally a study about the security and confidentiality of exchanged information.